\documentclass[10pt,twocolumn,letterpaper]{article}

\usepackage{cvpr}              

\usepackage[accsupp]{axessibility} 

\usepackage{xr}
\usepackage{epsfig}
\usepackage{graphicx,comment}
\usepackage{amsmath}

\usepackage{indentfirst}
\usepackage{algorithm}
\usepackage{algpseudocode} 
\usepackage{amssymb}
\usepackage{verbatim}
\usepackage{multirow}
\usepackage{booktabs}
\usepackage{longtable}
\usepackage{soul}
\usepackage{arydshln}
\usepackage{enumitem}
\usepackage[usenames,dvipsnames,table]{xcolor}
\usepackage{tikz}
\usepackage{float}
\usepackage{xspace}
\usepackage{bm}
\usepackage[titletoc]{appendix}
\usepackage[resetlabels]{multibib}
\usepackage{capt-of}

\usepackage{sidecap}

\newcites{sec}{References}
\newcommand{\redcrossbox}{
  \tikz[baseline=(X.base)]{
    \node[draw=black, fill=white, rectangle, minimum size=1em, inner sep=2pt] (X) {
      \textcolor{red}{\textbf{×}}
    };
  }
}

\newcommand{\SystemName}{QID}

\usepackage{upgreek}

\usepackage{color, colortbl}
\definecolor{codegreen}{rgb}{0,0.74,0.38}
\definecolor{codegray}{rgb}{0.5,0.5,0.5}
\definecolor{codepurple}{rgb}{0.58,0,0.82}
\definecolor{backcolour}{RGB}{230, 230, 230}
\definecolor{LightCyan}{rgb}{0.88,1,1}
\definecolor{LightRed}{RGB}{255, 204, 203}
\definecolor{LightGreen}{RGB}{204, 231, 207}

\usepackage{stackengine}

\usepackage{amsmath}


\makeatletter
\def\adl@drawiv#1#2#3{%
        \hskip.5\tabcolsep
        \xleaders#3{#2.5\@tempdimb #1{1}#2.5\@tempdimb}%
                #2\z@ plus1fil minus1fil\relax
        \hskip.5\tabcolsep}
\newcommand{\cdashlinelr}[1]{%
  \noalign{\vskip\aboverulesep
           \global\let\@dashdrawstore\adl@draw
           \global\let\adl@draw\adl@drawiv}
  \cdashline{#1}
  \noalign{\global\let\adl@draw\@dashdrawstore
           \vskip\belowrulesep}}
\makeatother

\usepackage{pifont}
\usepackage[pagebackref,breaklinks,colorlinks]{hyperref}
\hypersetup{
    colorlinks=true, 
    urlcolor=black,  
}
\usepackage[capitalize]{cleveref}
\crefname{section}{Sec.}{Secs.}
\Crefname{section}{Section}{Sections}
\Crefname{table}{Table}{Tables}
\crefname{table}{Tab.}{Tabs.}



\def\cvprPaperID{03} 
\def\confName{CVPR}
\def\confYear{2025}
\begin{document}
\title{QID: Efficient \underline{Q}uery-\underline{I}nformed ViTs in \underline{D}ata-Scarce Regimes \\ for OCR-free Visual Document Understanding}
\author{
    Binh M. Le\textsuperscript{1}\thanks{These authors contributed equally.} \thanks{Work done during internship at Amazon.} ,
    Shaoyuan Xu\textsuperscript{2}\footnotemark[1] ,
    Jinmiao Fu\textsuperscript{2},
    Zhishen Huang\textsuperscript{2},
    Moyan Li\textsuperscript{2},
    Yanhui Guo\textsuperscript{2}, \\
    Hongdong Li\textsuperscript{2,}\textsuperscript{3},
    Sameera Ramasinghe\textsuperscript{4},
    Bryan Wang\textsuperscript{2}\\
    \textsuperscript{1}Sungkyunkwan University, S. Korea\enspace
    \textsuperscript{2}Amazon, USA\enspace
    \textsuperscript{3}ANU, Australia\enspace
    \textsuperscript{4}Pluralis Research, Australia \\
    {\tt\small bmle@g.skku.edu \quad \{shaoyux,jinmiaof,hzs,moyanli,yanhuig,hongdli\}@amazon.com} \\
    {\tt\small samramasinghe@gmail.com \quad brywan@amazon.com}
}

\maketitle

\begin{abstract} 
In Visual Document Understanding (VDU) tasks, fine-tuning a pre-trained Vision-Language Model (VLM) with new datasets often falls short in optimizing the vision encoder to identify query-specific regions in text-rich document images. Existing methods that directly inject queries into model layers by modifying the network architecture often struggle to adapt to new datasets with limited annotations. To address this, we introduce QID, a novel, streamlined, architecture-preserving approach that integrates query embeddings into the vision encoder, leading to notable performance gains, particularly in data-scarce fine-tuning scenarios. Specifically, our approach introduces a dual-module framework: a query-aware module that generates a unique query vector to precisely guide the model's focus, as well as a query-agnostic module that captures the positional relationships among tokens, ensuring robust spatial understanding. Notably, both modules operate independently of the vision attention blocks, facilitating targeted learning of query embeddings and enhancing visual semantic identification. Experiments with OCR-free VLMs across multiple datasets demonstrate significant performance improvements using our method, especially in handling text-rich documents in data-scarce environments. 
\end{abstract} 
\section{Introduction}
\label{sec:introduction}
\begin{figure}[t]
\centering
\includegraphics[width=0.38\textwidth]{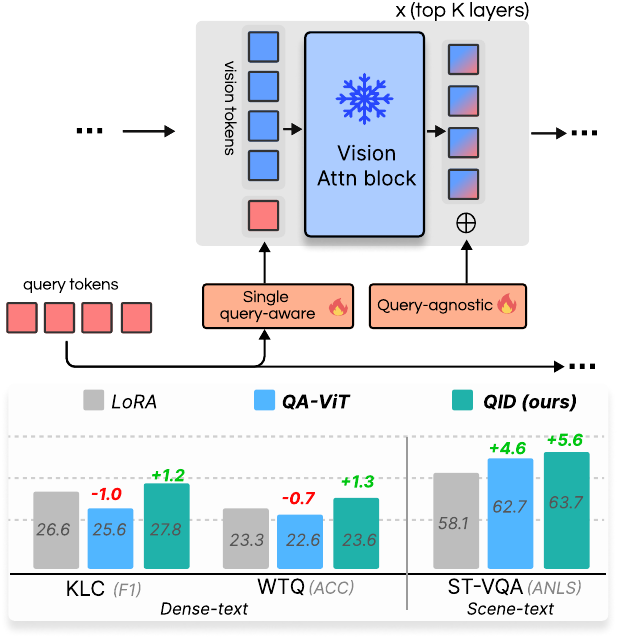}
\caption{\textbf{Illustration of our approach.} \textbf{Top:} Unlike previous work (e.g.,  QA-ViT \cite{ganz2024questionaware}), our method detaches the query-informed module \textit{independently} from the attention block, decomposing it to a query-aware module learning single embedding vector and a query-agnostic module, thereby reducing computational demands during training and inference. \textbf{Bottom}: Comparative results of our proposed method on two dense-text and one scene-text image datasets versus baseline methods, applying fine-tuning to the Qwen-VL-Chat model with only 1,000 samples per dataset.}
\label{fig:thumbnail_motivation}
\vspace{-12pt}
\end{figure}


Recent advancements in vision-language models (VLMs) have significantly impacted Visual Document Understanding (VDU), enabling models to interpret text-rich document images across various tasks. VDU methods are generally divided into two categories: optical character recognition (OCR)-dependent methods and OCR-free methods.  Although OCR-dependent methods \cite{appalaraju2021docformer, xu2020layoutlm, xu2020layoutlmv2, huang2022layoutlmv3, tang2023unifying, wang2023docllm} have obtained robust results, they are often bottlenecked by the computational latency and errors introduced by the OCR engines. \cite{kim2021donut, taghva2006effects}. In contrast, OCR-free approaches \cite{kim2021donut, ye2023ureader, ye2023mplugdoc, bai2023qwen, hu2024mplug} bypass these challenges, directly modeling the visual context of documents. These techniques generally utilize large VLMs trained in extensive VDU tasks \cite{masry2022chartqa, mathew2021docvqa, mishra2019ocrvqa, singh2019textvqa}, where visual and textual embeddings are jointly processed by language models to answer questions about an image.



However, fine-tuning pre-trained VLMs for VDU, especially with limited data, presents unique challenges. Unlike general visual question answering (VQA) tasks, VDU often requires a deeper contextual understanding and domain-specific knowledge, especially in dense text documents, where the annotation of logical inference can be complex \cite{mathew2022infographicvqa}. This challenge underscores the need for efficient adaptation in data-scarce settings, critical for VDU applications in fields like medical or legal document analysis, with minimal additional parameters. Existing Parameter-efficient fine-tuning (PEFT) techniques, such as LoRA \cite{hu2021lora} and DoRA \cite{liu2024dora}, are popular for adapting language models but struggle with high-complexity VDU tasks since they only enhance the linguistic components. More recently, a stream of research called query-aware (QA) \cite{dai2023instructblip, abramovich2024visfocus, ganz2024questionaware} has shown promise by injecting query embeddings into the visual encoding process, enhancing query-specific visual attention. However, these methods struggle with dense text documents under data-scarce conditions, where they introduce inefficiencies and fail to generalize effectively.

To address these limitations, we present \textbf{QID}, a novel, lightweight Query-Informed Vision Transformer (ViT) designed particularly in Data-scarce regimes. Our method introduces two modules: (1) a query-aware module, which generates a single, robust query embedding vector to align the model's focus with relevant document regions, and (2) a query-agnostic module, which captures positional dependencies across visual tokens and mitigates distribution shifts introduced by the query, to help the model maintain consistency in general layout patterns across diverse documents. To enhance local cross-attention, we also introduce fuse and defuse learning steps in the query-aware module, which refine query embeddings through spherical augmentation and entropy regularization. These steps enable precise query alignment with visual elements, even in data-limited scenarios. Both query-aware and query-agnostic modules operate independently of the core attention blocks, enabling efficient adaptation to query-relevant features with minimal data. As illustrated in Fig. \ref{fig:thumbnail_motivation}, our approach not only maintains architectural simplicity but also delivers substantial performance gains across OCR-free VDU tasks, particularly in data-scarce regimes. 

Our contributions are summarized as follows:
 \begin{itemize}
     \item We propose a lightweight, architecture-preserving approach that integrates query embeddings into the vision encoder without modifying the core attention blocks. This method enhances the VLM’s ability to focus on query-relevant regions in text-rich documents, particularly beneficial in data-scarce scenarios.
     
     \item Our framework introduces a query-aware module for augmenting question representation and a query-agnostic module to address positional dependencies and vision distribution shift. We further enhance the query-aware module with fuse and defuse learning steps, using spherical augmentation and entropy regularization to improve query alignment and robustness.

     \item Extensive experiments across multiple VDU datasets with pre-trained VLMs validate the efficiency and effectiveness of our approach, demonstrating consistent improvements over State-of-The-Art (SoTA) QA and PEFT methods in data-scarce settings.
      
 \end{itemize}

\section{Related Work}
\label{sec:relatedwork}
\subsection{Visual Document Understanding}
Visual Document Understanding (VDU) seeks to interpret and reason logically about a wide range of digitalized document images. Strategies in VDU are categorized into two main approaches. OCR-dependent approaches integrate images with an external OCR engine to annotate textual content, exemplified by works such as \cite{appalaraju2021docformer, xu2020layoutlm, huang2022layoutlmv3}. Alternatively, OCR-free approaches train large vision-language models (LVLMs) on extensive datasets, as seen in \cite{kim2021donut, bai2023qwen, ye2023mplugdoc}, enhancing capabilities without relying on OCR.
Notably, mPLUG-DocOwl \cite{ye2023mplugdoc} enhances the capabilities of LVLMs for VDU by introducing a modular model based on mPLUG-Owl \cite{ye2023mplug} designed specifically for OCR-free document understanding. Pix2Struct \cite{lee2023pix2struct} pioneers an approach of screenshot parsing objectives, and UReader \cite{ye2023ureader} innovates with a shape-adaptive cropping module that precedes the encoder-decoder architecture, leveraging a frozen low-resolution vision encoder for processing high-resolution images.  However, a common limitation of these OCR-free approaches is that the vision encoder processes images without contextual knowledge of the textual prompts, which can lead to suboptimal vision representation for VDU tasks.

\subsection{Query-Informed ViTs}
InstructBLIP \cite{dai2023instructblip} attempted early textual instruction integration into vision embeddings via a QFormer atop the vision encoder, potentially overlooking image representation nuances. Subsequently, VisFocus \cite{abramovich2024visfocus} was proposed to encode prompts and perform cross-attention with vision tokens at every layer, but its performance did not surpass that of baseline models, as it was targeted at developing a lightweight model.  Recently,  QA-ViT \cite{ganz2024questionaware} advocates the idea of appending the prompt tokens directly to the vision tokens in some of the final vision attention blocks. While this method has shown promise in simple VQA tasks, it necessitates modifications to the original vision blocks, resulting in a much-increased number of training parameters.  Furthermore, when fine-tuning with dense-text datasets, applying the QA-based methods on top of LoRA \cite{hu2021lora} degrades the performance considerably, compared to using LoRA alone, especially in data-scarce regimes. 

\begin{figure*}[t!]
\centering
\includegraphics[width=0.86\textwidth]{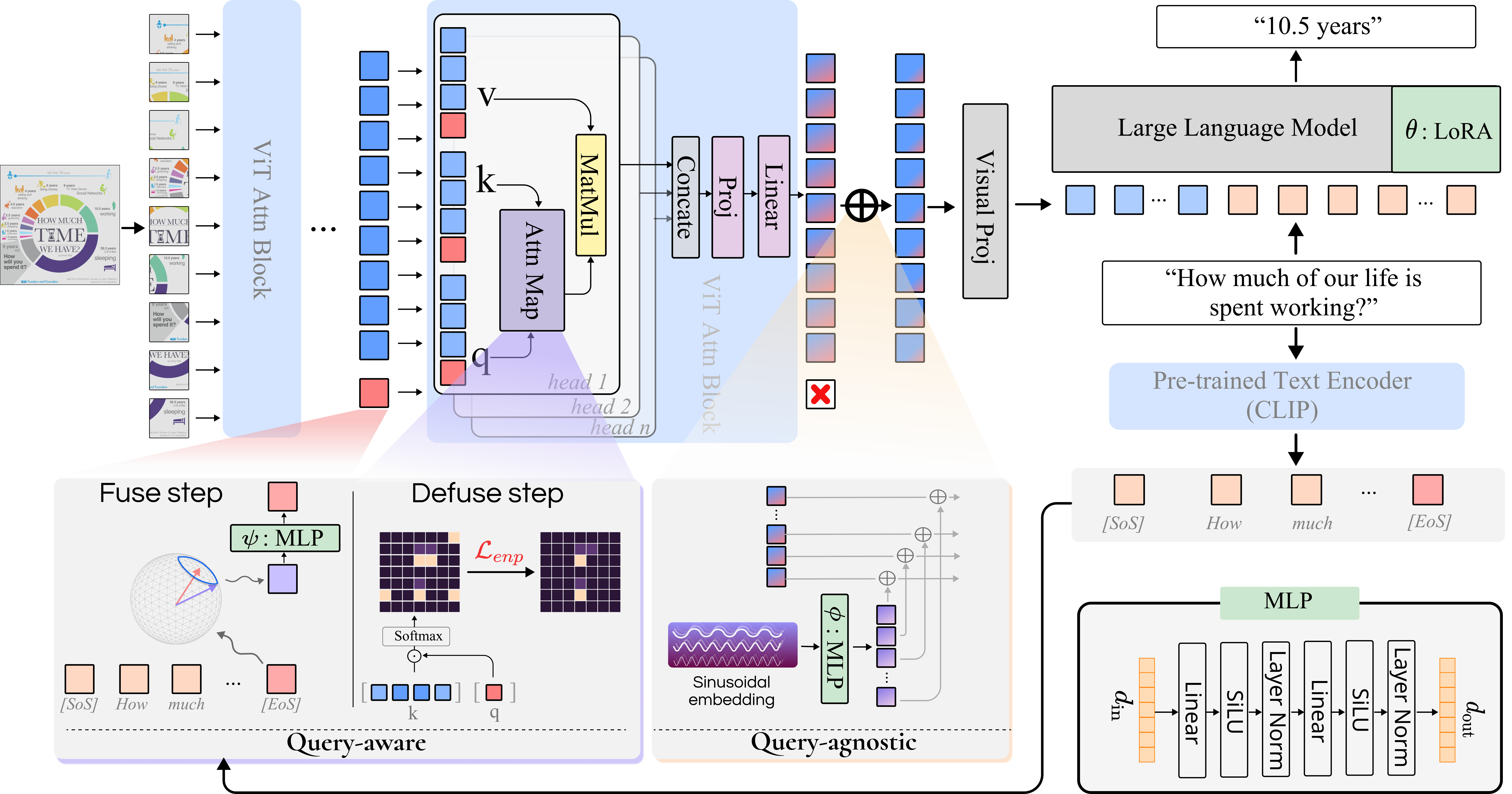}
\caption{ \textbf{Illustration of our end-to-end training procedure.} For simplicity, this figure demonstrates how our approach is integrated with the last ViT attention block. Note it can also be applied to other layers of an vision encoder.  During fine-tuning stage, only the \colorbox{LightGreen}{green modules} are optimized. Our query-aware module, enhanced by the fuse and defuse learning steps, makes the query embedding more robust for the vision encoder. Our query-agnostic module offsets distribution shifts caused by the query information and, as it operates independently from the query vector, can be precomputed and saved as a bias term post-training. This efficient learning approach on a single query vector makes our proposed method lightweight and highly effective for VLMs in VDU tasks.}
\label{fig:main_diagram}
\vspace{-14pt}
\end{figure*}

\section{Methods}
\label{sec:method}

\subsection{Overall Architecture}

 Figure \ref{fig:main_diagram} gives an overview of our method, showing how it integrates the query embedding into the vision attention block. Unlike previous methods\cite{ganz2024questionaware, abramovich2024visfocus}, we adopt the [\texttt{EoS}] (end-of-sentence) token of a query embedding from a pre-trained text encoder and make no modification to the network architecture of the vision attention block. This token vector is initially processed through a query-aware module, which injects the question embedding into the vision block. This query-aware module consists of two steps, i.e., fuse and defuse steps, aiming to enhance the query embedding under data-scarce conditions. The fuse step augments the query embedding on a hypersphere, while the defuse step focuses its attention on the most relevant visual areas. The vision tokens are subsequently adjusted by a query-agnostic module that learns a sinusoidal embedding and adds it directly to the vision tokens. Such query-agnostic components can be pre-computed after training, and used directly during inference to save computation. 

The notations used in this paper are defined as follows: Given an image $\mathcal{I}$ and a query (i.e. the question) $\mathcal{Q}$, the vision encoding output from a pre-trained vision encoder ${V}$, which consists of $L$ layers and is conditionally dependent on $\mathcal{Q}$, can be formalized as follows: 
\begin{equation}
\textbf{z}^L = \{z_i^L\}_{i=1}^{T_v} = {V}(\mathcal{I} \mid \mathcal{Q}), \quad z_i^L \in \mathbb{R}^{d_v},\end{equation} 
where $T_v$ and ${d_v}$ represent the number and the dimension of vision tokens, respectively. Typically, ${V}$ is represented as a stack of $L$ identical attention blocks, where each block is defined by: 
\vspace{-8pt}
\begin{align}
    \bar{\textbf{z}}^{l} &= {\texttt{Proj}}(\texttt{MSA}(\textbf{z}^{l-1}))+\textbf{z}^{l-1} \\
    \textbf{z}^{l} &= \texttt{FFN}(\bar{\textbf{z}}^{l}) + \bar{\textbf{z}}^{l},
\end{align}
where \texttt{MSA}, \texttt{Proj}, and \texttt{FFN} denote the multi-head self-attention, projection, and feed forward multi-layer perceptron layers, respectively. 
 
\subsection{Query-Aware Module}
\subsubsection{Single Query Embedding}

With the increasing prevalence of decoder-only large language models, such as GPT-3 \cite{brown2020gpt} and LLaMA \cite{touvron2023llama}, encoding a query (or question) often requires an external pre-trained text encoder.  In this paper, we employ the text encoder from CLIP \cite{radford2021learning}, known for its effectiveness in integration with multi-modal models for generation tasks \cite{kim2022diffusionclip} and segmentation tasks \cite{bousselham2024grounding}.

\begin{figure}[t!]
\centering
\includegraphics[width=0.40\textwidth]{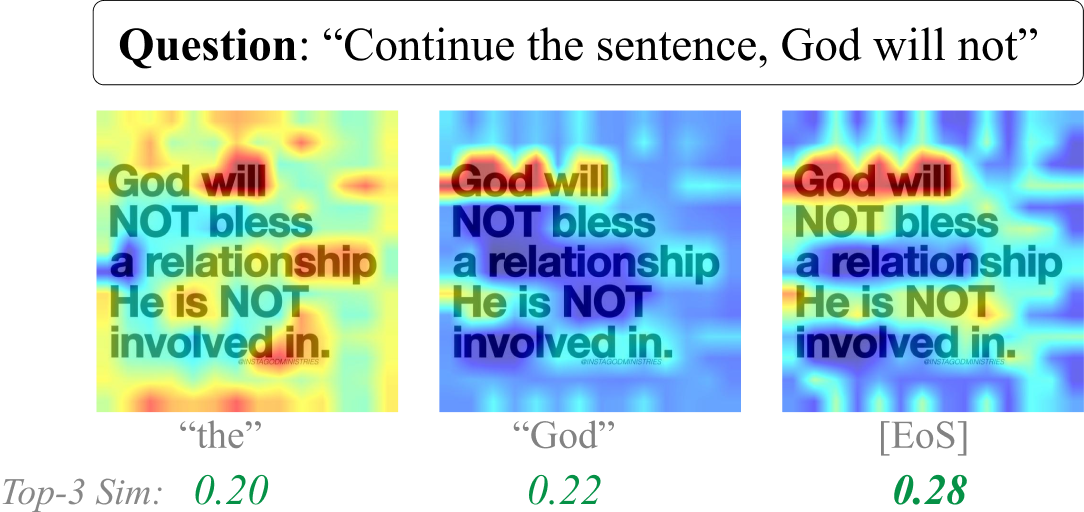}
\caption{Effect of the {[\texttt{EoS}]} token in the query for highlighting semantic areas in the image \cite{li2023clip}. The green numbers at the bottom indicate the top-3 cosine similarities between tokens and the image, as computed using CLIP embeddings \cite{radford2021learning}.}
\label{fig:eos_token}
\vspace{-14pt}
\end{figure}

Let $\textbf{q} \in \mathbb{R}^{d_t \times T_t}$ represent the text embedding of a query generated by the pre-trained text encoder, where $d_t$ is the dimension of the text embedding space, and $T_t$ is the maximum number of text tokens. While previous work \cite{ganz2024questionaware} uses full text embeddings to train the model to capture the entire query representation, we find this approach suboptimal in data-scarce settings. Furthermore, \cite{li2023clip} suggests that in models like CLIP, the end-of-sentence ([\texttt{EoS}]) token, punctuation marks, and non-object words often highlight corresponding semantic regions in images, whereas other tokens typically do not. Our findings confirm this, as illustrated in Fig. \ref{fig:eos_token} 
, where the top three text tokens most similar to the image in the CLIP model \cite{radford2021learning} reveal that [\texttt{EoS}] highlights the most relevant visual area, while tokens with lower similarity, such as ``the," may even highlight background regions. This outcome arises from the fact that the [\texttt{EoS}] token serves as a holistic sentence representation in contrastive learning with vision tokens within the CLIP model.  Therefore, using a single [\texttt{EoS}] token, denoted as $q_{\textrm{eos}} \in \mathbb{R}^{d_t}$, from the query embedding offers two key benefits. First, it is the most efficient token for highlighting important spatial areas in an image. Second, with limited fine-tuning data, optimizing the downstream module to integrate query information into the vision model using a single token is both more efficient and robust than using all text tokens, which often introduce additional noise.

\subsubsection{Fuse and Defuse Learning Steps}

Through experiments, we have found that combining  QA-ViT  \cite{ganz2024questionaware} with PEFT methods, such as LoRA \cite{hu2021lora}, often yields inferior results compared to using LoRA alone, particularly with limited fine-tuning data.  For example, as shown in Table \ref{tab:qa_single} and further detailed in Section \ref{sec:results},  our experiments with the Qwen-VL-Chat model on 1,000 fine-tuning samples show degraded performance using a single token,  QA-ViT+$q_{\textrm{eos}}$, compared with that with full query token,  QA-ViT+$\textbf{q}$. 
{We hypothesize that as  the  QA-ViT  method \cite{ganz2024questionaware} introduces a new \texttt{Proj} layer into the ViT block, it modifies the pre-trained model's attention dynamics without sufficient samples to optimize the new parameters, thereby impairing visual representation quality.}


\begin{table}[t!]
\centering
\resizebox{0.41\textwidth}{!}{ 
\begin{tabular}{l|ccc|cc}
\hline 

    \textbf{Methods}   &  \textbf{InfoVQA} & \textbf{KLC} & \textbf{WTQ} & \textbf{VizWiz} & \textbf{ST-VQA}    \\ 
    &\textcolor{gray}{\textit{ANLS $\uparrow$}} & \textcolor{gray}{\textit{F1 $\uparrow$}} & \textcolor{gray}{\textit{ACC $\uparrow$}} & \textcolor{gray}{\textit{VQA Score $\uparrow$}} & \textcolor{gray}{\textit{ANLS $\uparrow$}} \\
    \hline\hline 
\multicolumn{6}{c}{1K fine-tuning samples}  \\ 
\hline 
LoRA &34.38	&26.62	&23.30		&38.46	&58.07\\

 QA-ViT +$\textbf{q}$        & 33.80   & 25.64   & 22.08   & 40.40   & 62.73                     \\
 QA-ViT +$q_{\textrm{eos}}$ & 34.01   & 25.96   & 21.69   & 38.55   & 61.58                     \\
\SystemName\ (\textit{ours}) &   {34.18} &   {27.81} &   {23.56} &   {42.08} &   {63.69} \\
\hline \hline 
\end{tabular}}
\caption{Comparison between the baseline method  QA-ViT  + $q_{\textrm{eos}}$ and our approach, \SystemName\ + 
 $q_{\textrm{eos}}$, as experimented with the Qwen-VL-Chat model fine-tuned with 1,000 samples. Experimental settings are  further detailed in Section \ref{sec:results}.}
\label{tab:qa_single}
\vspace{-13pt}
\end{table}

  Building on the observed shortcomings of QA-ViT  \cite{ganz2024questionaware}, we introduce a training paradigm that preserves the original architecture of the ViT model's attention block while utilizing a single token embedding enhanced by ``fuse" and ``defuse" learning steps, which are illustrated in Fig. \ref{fig:main_diagram}. The ``fuse" step enriches the query embeddings by augmenting them with a random noise vector from a Gaussian distribution around the original vector. On the other hand, the ``defuse" step eliminates unrelated visual information (noises) while retaining the most relevant elements through entropy regularization.

\textbf{Fuse step: spherical augmentation.}  Pre-trained CLIP models \cite{radford2021learning} are trained by maximizing the cosine similarity of text and image features for matching text-image pairs while minimizing it for mismatched pairs. 
Despite their extensive training on large datasets, there is still a gap remaining between image embeddings and corresponding text embeddings, rooting from the cone effect inherent in each modality's distribution \cite{liang2022mind}.  
Moreover, in our use case, where there is no image caption in the input but only a question about it, relying on separately pre-trained text and image encoders may underperform for VDU tasks. These observations motivate us to explore the potential of embedding space of $q_{\textrm{eos}}$ to alleviate the narrow cone distribution and enhance the semantic representation of question embeddings. Meanwhile, this augmentation step fosters robust learning in scenarios with limited fine-tuning data by encouraging it to learn more generalized and resilient visual representations with respect to the query. Formally, we form pseudo text features $q_{\textrm{eos}}^{\prime} \in \mathcal{S}(\mathcal{Q})$ for a given question $\mathcal{Q}$ on the hypersphere:
\begin{equation}
    \mathcal{S}(\mathcal{Q}) = \{ q_{\textrm{eos}}^{\prime} | \textrm{Sim}(q_{\textrm{eos}}^{\prime},  q_{\textrm{eos}}) > \tau \},
\end{equation}
where $\textrm{Sim}$ denotes cosine similarity and $\tau$ is a threshold. 

To generate a pseudo query feature $q_{\textrm{eos}}^{\prime}$, we introduce a method to perturb the original query feature $q_{\textrm{eos}}$ using adaptive Gaussian noise:
\begin{equation}
     q_{\textrm{eos}}^{\prime} = \frac{\widetilde{{q}}_{\textrm{eos}}}{\| \widetilde{{q}}_{\textrm{eos}}\|_2}, \quad \widetilde{{q}}_{\textrm{eos}} = q_{\textrm{eos}} + \sigma \cdot  \| q_{\textrm{eos}}\|_2 \cdot \frac{\epsilon}{\|\epsilon\|_2},
\end{equation}
where $\epsilon \sim \mathcal{N}(\textbf{0}, \textbf{I})$ represents Gaussian noise, $\sigma > 0$ is the hyperparameter that dictates the perturbation magnitude, and $\|\cdot\|_2$ denotes the $\ell_2$ norm. The normalization of the Gaussian noise onto a hypersphere, followed by rescaling according to the norm of the query feature, ensures that the noise addition is adaptive. Subsequently, $q_{\textrm{eos}}^{\prime}$ is processed through a multi-layer perceptron (MLP) $\psi$ to project it into the vision space $\mathbb{R}^{d_v}$. With the output $\textbf{z}^{l-1}$ at the layer $(l-1)$-th in the vision model, the input for the next layer $l$-th is formed as $\textbf{z}^{l-1} = \text{Cat}\{\textbf{z}^{l-1} , \psi(q_{\textrm{eos}}^{\prime})\} \in \mathbb{R}^{(T_v+1)\times d_v}$, where $\text{Cat}$ denotes a simple concatenation operation. This step in the process is illustrated in Fig. \ref{fig:main_diagram} - bottom left.

\textbf{Defuse step: entropy regularization.} In text-centric images, relevant visual areas are usually confined to small, localized regions (as  illustrated in Fig. \ref{fig:qualitative_result}). Therefore, this step aims to constrain the query embedding to activate only specific local vision tokens via their cross-attention matrices. In the $l$-th vision attention block, the multi-head self-attention (\texttt{MSA}) layer consists of $H$ distinct heads. For each head $h$ ($1 \leq h \leq H$), the token embedding ${z}^{l-1}_i$ from the previous layer is projected into triplet forms: query, key, and value (note that this query is different from the query referring to the original question $\mathcal{Q}$). The matrices for query ($Q^l_h$), key ($K^l_h$), and value ($V^l_h$) contain corresponding elements. The self-attention matrix for the $h$-th head is:
\begin{equation}
    A^l_h = \text{softmax}\left({Q^l_h(K^l_h)^T}/{\sqrt{d_v}}\right) \in \mathbb{R}^{(T_v+1)^2},
\end{equation}
where $\text{softmax}$ is applied across the columns of the inner dot product. The cross-attention between the $q_{\textrm{eos}}^{\prime}$ question embedding and the vision tokens is defined as:
\begin{equation}
    A_{h| \textrm{cross}}^l = A^l_h[T_v+1, :T_v] \in \mathbb{R}^{T_v}.
\end{equation}
We measure the uncertainty of a distribution using entropy over $A_{h| \textrm{cross}}^l$:
\vspace{-10pt}
\begin{equation}
    \mathcal{H}_{h}^l = -\sum_{i=1}^{T_v} (A_{h| \textrm{cross}}^l)_i \cdot \log\big[ (A_{h| \textrm{cross}}^l)_i \big].
\end{equation}
To discourage a uniform distribution of $A_{h| \textrm{cross}}^l$ over the spatial dimension, we apply entropy regularization as follows:
\vspace{-8pt}
\begin{equation}
    \mathcal{L}_\text{enp} = \frac{1}{|L_{q}|} \sum_{l \in L_{q}} \sum_{h=1}^{H} \mathcal{H}_{h}^l,
\end{equation}
where $L_\text{q}$ are the predefined visual layers to which we project the query embedding $q_{\textrm{eos}}^{\prime}$. The end-to-end of this process is depicted in Fig. \ref{fig:main_diagram}.

\subsection{Query-Agnostic Module}
To offset the distribution shift caused by the introduction of an additional query token $q_{\textrm{eos}}^{\prime}$ to the frozen vision attention block, we implemented a query-agnostic module. This module functions as a bias term added to the output of the block. Unlike QA-ViT  \cite{ganz2024questionaware}, which modifies the attention block based on the query token, our module operates independently, learning solely from an initiated sinusoidal signal and is inserted directly after the vision block as illustrated in Fig. \ref{fig:main_diagram}. Formally, a fixed position embedding of dimension $d_p$ employs a sinusoidal function \cite{vaswani2017attention}:
{\small \begin{equation}
    \textbf{P}[i,2k] = \sin\left( \frac{i}{10000^\frac{2k}{d_p}} \right), 
    \textbf{P}[i,2k+1] = \cos\left( \frac{i}{10000^\frac{2k}{d_p}} \right).
\end{equation}} 
Here, $\textbf{P} \in \mathbb{R}^{T_v \times d_p}$, $i$ denotes the index of the position, and $k$ represents the index within the dimension of the embedding. We set the value of $d_p$ to 64 in our work. Each spatial sinusoidal vector is further refined by a learnable, shallow feed-forward multi-layer perceptron (MLP) $\phi$, and matched in dimension to the vision space $\mathbb{R}^{d_v}$. The output from the vision attention block $\mathbf{z}^{l}$ that is processed through a query-aware module. After removing the query token \cite{ganz2024questionaware} (\redcrossbox in Fig. \ref{fig:main_diagram}), this output is enhanced by adding the learnable embedding:
\begin{equation}
    \textbf{z}^{l} = \textbf{z}^{l} + \phi(\textbf{P}) \in \mathbb{R}^{T_v \times d_v}.
\end{equation}
Adding this learnable embedding serves dual purposes. Firstly, it compensates for the distribution shift, as mentioned earlier. Secondly, the sinusoidal embedding enhances spatial position awareness within the vision model \cite{lei2024scaffolding, dorkenwald2024pin}, which is beneficial for relative position localization tasks, such as table question-answering in the WTQ dataset \cite{pasupat2015compositional}.\\

\subsection{Training Objectives}

Let $\theta$ represent  optional fine-tune parameters of the LLM part (e.g., LoRA parameters  \cite{hu2021lora}). Our training objective for efficiently injecting the query token into the vision model of the VLM is formulated as:
\vspace{-8pt}
\begin{equation}
    \min_{\psi, \phi, \theta} \mathcal{L}_\text{Overall} =\mathcal{L}_\text{CE} + \alpha \mathcal{L}_\text{enp},
\end{equation}
where $\mathcal{L}_\text{CE}$ is the negative log-likelihood loss for the predicted text tokens of the fine-tuning dataset, and $\alpha$ is a hyper-parameter balancing the effect of entropy regularization on the vision attention blocks.

During inference, neither the fuse nor defuse steps is employed. Instead, only the single query $q_{\textrm{eos}}$ is used as $\psi(q_{\textrm{eos}})$.  Since the query-agnostic module is independent of the query tokens, after training, the matrix $\textbf{P}$ is processed through $\phi$ and saved as a bias matrix, significantly reducing computational overhead.

\section{Experiments}
\label{sec:results}

\subsection{Settings}

\textbf{VLM models.} To demonstrate the generalization of our proposed method, we selected two SoTA large VLMs: mPLUG-Owl2 \cite{ye2024mplugowl2} and Qwen-VL-Chat \cite{bai2023qwen}. Both models have approximately 7 billion parameters. The former model, mPLUG-Owl2, was trained on various VQA datasets and has shown SoTA performance on OCR tasks. The latter, Qwen-VL-Chat, in addition to being trained with VQA datasets, has been trained with a variety of text-centric datasets such as Doc-VQA \cite{mathew2021docvqa}, TextVQA \cite{singh2019textvqa}, OCRVQA \cite{mishra2019ocrvqa}, and ChartQA \cite{masry2022chartqa}.
\begin{table}[b!]
\vspace{-10pt}
\centering
 \resizebox{0.45\textwidth}{!}{ 
\begin{tabular}{l|l|c|c|c}
\hline
\multicolumn{2}{c|}{\textbf{Dataset}}       & \textbf{Task}  & \textbf{Training Set} & \textbf{Test Set} \\
\hline \hline
\multirow{3}{*}{{\parbox[c]{1cm}{\centering Dense text}}} &InfoVQA   \cite{mathew2022infographicvqa}    & VQA   & 24K & 3.3K     \\
&KLC  \cite{stanislawek2021kleister} & KIE   & 14K & 4.9K     \\
&WTQ  \cite{pasupat2015compositional} & Table & 14K & 4.3K     \\ \hline
\multirow{2}{*}{{\parbox[c]{1cm}{\centering Scence text}}} &VizWiz \cite{gurari2018vizwiz} & VQA   & 21K & 4.3K     \\
&ST-VQA \cite{biten2019scene} & VQA   & 20K & 6K      \\
\hline \hline
\end{tabular}
}
\caption{ Statistics of the fine-tuning and evaluation datasets. For ST-VQA dataset, we split its public training set into training and test set. As per the experimental settings outlined in Table \ref{tab:main_results}, only a subset of each dataset (e.g., 1,000 training samples) is utilized for fine-tuning, while test sets are hold the same.}
\label{tab:dataset}
\end{table}
\begin{table*}[th!]
\centering
 \resizebox{0.82\textwidth}{!}{ 
    \begin{tabular}{l|c|l|c|ccc|cc|c}
    \hline 
    \multirow{2}{*}{\textbf{Models}} & \multirow{2}{*}{\textbf{Resolution}} & &\textbf{+Ovh./ViTBlock} & \textbf{InfoVQA} & \textbf{KLC} & \textbf{WTQ} & \textbf{VizWiz} & \textbf{ST-VQA} & \textbf{\textit{Avg.}} \\
    &&  & \textcolor{gray}{\textit{GFLOP $\downarrow$}} &\textcolor{gray}{\textit{ANLS $\uparrow$}} & \textcolor{gray}{\textit{F1 $\uparrow$}} & \textcolor{gray}{\textit{ACC $\uparrow$}} & \textcolor{gray}{\textit{VQA Score $\uparrow$}} & \textcolor{gray}{\textit{ANLS $\uparrow$}} \\
    \hline \hline
    {\textcolor{gray}{UREADER \cite{ye2023ureader} }} & \textcolor{gray}{{($896\times 896$)}} &\multirow{7}{*}{\textcolor{gray}{\textit{Fully trained}}} & & \textcolor{gray}{42.2} & \textcolor{gray}{32.8} & \textcolor{gray}{29.4} & - &- &-\\ 
    {\textcolor{gray}{Pix2Struct-L \cite{lee2023pix2struct} }} & \textcolor{gray}{{($1024\times 1024$)}} & & & \textcolor{gray}{40.0} & - & -& - &- &- \\ 
    {\textcolor{gray}{Pali-3 \cite{chen2023pali3} }} & \textcolor{gray}{{($1064\times 1064$)}} & & & \textcolor{gray}{57.8} & - & -& - &- &-\\ 
    {\textcolor{gray}{Qwen-VL-Chat$^\dagger$ \cite{bai2023qwen} }} & \textcolor{gray}{{($448\times 448$)}} & & &\textcolor{gray}{33.1} & \textcolor{gray}{31.5} & \textcolor{gray}{24.8}  & - &- &-\\ 
    {\textcolor{gray}{mPLUG-Owl$^\dagger$\cite{ye2023mplug}}} & \textcolor{gray}{{($448\times 448$)}} & & & \textcolor{gray}{32.5} & \textcolor{gray}{31.2} & \textcolor{gray}{25.2}& - &- &-\\ 
    {\textcolor{gray}{mPLUG-DocOwl\cite{ye2023mplugdoc}}} & \textcolor{gray}{{-}} & & & \textcolor{gray}{38.2} & \textcolor{gray}{30.3} & \textcolor{gray}{26.9}& - &- &-\\ 
    {\textcolor{gray}{Visfocus \cite{abramovich2024visfocus}}} & {-} & & &\textcolor{gray}{31.9} & - & -& - &- &-\\ 
    {\textcolor{gray}{LLaVA+Vicuna+QA-ViT \cite{ganz2024questionaware}}} & \textcolor{gray}{{($336\times 336$)}} & & & - & - & -& - &\textcolor{gray}{62.4} &-\\ 
\hline
    \multirow{13}{*}{mPLUG-Owl2 \cite{ye2024mplugowl2} } &  \multirow{13}{*}{{($448\times 448$)}}& \textit{Zero-shot}  & & 25.0   & 9.4    & 11.9   & 32.5   & 55.0   & 26.8                 \\

    \cline{3-10} 
    && \textit{\textbf{PEFT on 1K}}   & & & & & & &  \\
    && \quad DoRA \cite{liu2024dora} & & 25.6   &  {16.7}   & 12.2   & 50.4   & 54.9   & 32.0                 \\
    && \quad LoRA \cite{hu2021lora} & & 25.5   & 15.7   &  {12.6}   &  {51.5}   & 54.7   &  {32.0}                 \\
    && \quad VPT \cite{jia2022visual} & +0.84 & 25.8   & 16.1   & 11.6   & 50.1   & 53.5   & 31.4                 \\
    && \quad  QA-ViT  \cite{ganz2024questionaware} & +1.88 & \textbf{26.7}   & 16.5   & 12.6   & 47.1   &  {55.2}   & 31.6                 \\

    && \cellcolor{backcolour}\quad \SystemName\ (\textit{\textbf{ours}}) & \cellcolor{backcolour}{\textbf{+0.009}} & \cellcolor{backcolour}{\textbf{26.7}} & \cellcolor{backcolour}{\textbf{16.8}} & \cellcolor{backcolour}{\textbf{12.8}} & \cellcolor{backcolour}{\textbf{52.0}} & \cellcolor{backcolour}{\textbf{55.4}} & \cellcolor{backcolour}{\textbf{32.7}}        \\

\cline{3-10} 
&& \textit{\textbf{PEFT on 2K}}  & & & & & & & \\
    && \quad DoRA \cite{liu2024dora} & & 25.9   & 16.4   &  {12.9}   & 51.6   & 55.2   & 32.4       \\
    && \quad LoRA \cite{hu2021lora} & &  {26.3}   & 16.5   & 13.1   &  {52.7}   & 55.3   &  {32.8}     \\
    && \quad VPT \cite{jia2022visual} & +0.84 & 26.0   &  {17.1}   & 12.1   & 51.3   & 54.6   & 32.2       \\
    && \quad  QA-ViT  \cite{ganz2024questionaware} &  +1.88 & 25.6   & 16.7   & 12.6   & 52.5   &  {55.4}   & 32.6     \\
    && \cellcolor{backcolour}\quad \SystemName\ (\textit{\textbf{ours}}) & \cellcolor{backcolour}{\textbf{+0.009}} & \cellcolor{backcolour}{\textbf{27.3}} & \cellcolor{backcolour}{\textbf{17.4}} & \cellcolor{backcolour}{\textbf{13.4}} & \cellcolor{backcolour}{\textbf{53.9}} & \cellcolor{backcolour}{\textbf{55.7}} & \cellcolor{backcolour}{\textbf{33.5}}
\\
    \hline
    \multirow{13}{*}{Qwen-VL-Chat \cite{bai2023qwen} } &  \multirow{13}{*}{{($448\times 448$)}}   & \textit{Zero-shot} & &34.2   & 19.9   & 22.6   & 35.2   & 56.5   & 33.7  
\\
    \cline{3-10} 
    && \textit{\textbf{PEFT on 1K}}  & & & & & & &  \\
    && \quad DoRA \cite{liu2024dora} & & \textbf{34.6} & 27.3   & 23.1   & 39.0   & 58.1   & 36.4 \\
    && \quad LoRA \cite{hu2021lora} & & 34.4   & 26.6   & 23.3   & 38.5   & 58.1   & 36.2        \\
    && \quad VPT \cite{jia2022visual} & +2.01 & 31.0   & 26.0   & 20.6   & \textbf{43.3} & 56.2   & 35.4   \\
    && \quad  QA-ViT \cite{ganz2024questionaware} &  +4.53 & 33.8   & 25.6   & 22.6   &  40.4 & 62.7  & 36.9   \\
    && \cellcolor{backcolour}\quad \SystemName\ (\textit{\textbf{ours}}) & \cellcolor{backcolour}{\textbf{+0.02}} & \cellcolor{backcolour}34.2   & \cellcolor{backcolour}\textbf{27.8} & \cellcolor{backcolour}\textbf{23.6} & \cellcolor{backcolour} {42.1}    & \cellcolor{backcolour}\textbf{63.7} & \cellcolor{backcolour}\textbf{38.3}
\\
\cline{3-10} 
&& \textit{\textbf{PEFT on 2K}} & & & & & & & \\
    && \quad DoRA \cite{liu2024dora} & & 34.5 & 27.8 & 23.4 & 39.2       & 58.0 & 36.6 \\
    && \quad LoRA \cite{hu2021lora} & & 34.4 & 27.4 & 23.2 & 38.9       & 58.0 & 36.4 \\
    && \quad VPT \cite{jia2022visual} & +2.01 & 31.9 & 26.9 & 20.8 & \textbf{43.4}       & 56.8 & 36.0\\
    && \quad  QA-ViT  \cite{ganz2024questionaware} & +4.53 & 34.5 & 28.4 & 24.5 & 42.0       & 64.2 & 38.7 \\
    && \cellcolor{backcolour}\quad \SystemName\ (\textit{\textbf{ours}}) & \cellcolor{backcolour}{\textbf{+0.02}} &  \cellcolor{backcolour}\textbf{34.8} & \cellcolor{backcolour}\textbf{28.8} & \cellcolor{backcolour}\textbf{24.9} & \cellcolor{backcolour} {42.2} & \cellcolor{backcolour} \textbf{65.1} & \cellcolor{backcolour}\textbf{39.1}
\\
    \hline \hline
    \end{tabular}%
}
\caption{\textbf{Experimental results across five datasets}.  \textit{Ovh./ViTBlock} denotes the computational overhead added to each Vision Transformer (ViT) block during the inference phase. $\dagger$ denotes models trained using the approach described in \cite{li2024enhancing}.  $\downarrow$ and $\uparrow$ indicate that lower and higher values are preferable, respectively. Our method offers a substantial improvement with modest overhead, with greater impact when dataset annotations require expert knowledge.}
\label{tab:main_results}
\end{table*}

\textbf{Datasets and evaluation metrics.} During the fine-tuning phase, we intentionally excluded datasets that were used in the pre-training phase (e.g., DocVQA, TextVQA, OCR-VQA, ChartQA) to avoid redundancy and ensure the robustness of the fine-tuning process. Consequently, the two VLMs, mPLUG-Owl2 and Qwen-VL-Chat, were fine-tuned with different text-centric datasets, including both dense text datasets and scene text datasets as outlined in Table \ref{tab:dataset}. The dense text datasets, which primarily feature images captured from documents, include InfoVQA \cite{mathew2022infographicvqa}, KLC \cite{stanislawek2021kleister}, and WTQ\cite{pasupat2015compositional}. Scene text datasets feature images from natural scenes and include VizWiz \cite{gurari2018vizwiz}  and ST-VQA  \cite{biten2019scene}.  More details of the datasets are provided in Supplementary Material - Section \ref{supp:supp_data}.

In alignment with previous studies \cite{ye2023mplugdoc, abramovich2024visfocus, bai2023qwen, li2024enhancing}, we report the performance of the InfoVQA and ST-VQA datasets using the Average Normalized Levenshtein Similarity (ANLS). For the KLC dataset, we use the $F_1$ score as the evaluation metric. The evaluation metric for WTQ is accuracy, while the VizWiz dataset is evaluated using the VQA score.
\begin{figure*}[t!]
\centering
\includegraphics[width=0.84\textwidth]{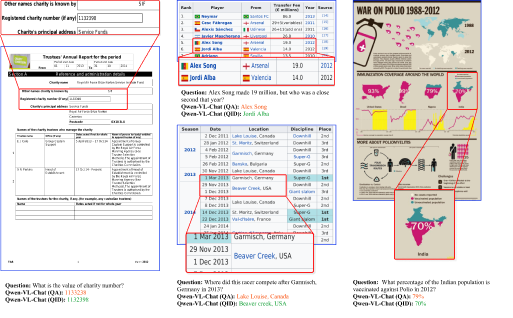}
\caption{ \textbf{Qualitative results between  QA-ViT  and our \SystemName}. Crucial regions are enlarged for better visualization. More visualizations are provided in Supp. Material - Section \ref{supp:Visualization}.}
\label{fig:qualitative_result}
\vspace{-10pt}
\end{figure*}

\textbf{Baselines.} To fine-tune the VLMs on target datasets, Parameter-Efficient Fine-Tuning (PEFT) methods are preferred. We compare our methods with LoRA \cite{hu2021lora} and its advanced version, DoRA \cite{liu2024dora}. Additionally, we have re-implemented Visual Prompt Tuning (VPT) employing 50 tokens on the vision encoders, and re-implemented QA-ViT  \cite{ganz2024questionaware} on both VLMs. 

\textbf{Implementation details.} During the fine-tuning stages, all images are resized to $448\times 448$ pixels. We set the values of $\alpha$ and $\sigma$ to $10^{-2}$ and $0.16$, respectively, by carefully tuning them to achieve an optimal balance. We note that setting them too high can result in either excessive noise (in case of $\sigma$) or overly aggressive filtering of visual information (in case of $\alpha$)). Experiments are conducted on a computational platform equipped with four NVIDIA A100 40GB GPUs. The models undergo the fine-tuning process with a cumulative batch size of 128, targeting a maximum of five epochs for both Qwen-VL-Chat and mPLUG-Owl2 models. Early stopping is applied when there is no decrease in validation loss. The training employs the AdamW optimizer \cite{loshchilov2017adamw}, starting with a learning rate of $2 \times 10^{-5}$ and a linear warm-up over 100 steps. 

For the main experimental results, we fine-tuned two models using a small number of samples from each dataset, specifically, 1,000 and 2,000 samples. In the ablation studies, we demonstrate that our method remains superior given very limited or full datasets.
\subsection{Comparisons with SoTA}
The performance outcomes of various VLMs on different benchmarks related to visual document understanding are consolidated in Table \ref{tab:main_results}. Specifically, for both models, mPLUG-Owl2 and Qwen-VL-Chat, our proposed method achieves the highest performance on average while introducing the smallest additional computational overhead on the Vision Transformer block, at $9 \times 10^{-3}$ GFLOPs for mPLUG-Owl2 and $2 \times 10^{-2}$ GFLOPs for Qwen-VL-Chat, respectively. With mPLUG-Owl2, we observe up to 0.7\% and 0.7\% improvements on average in the 1,000 and 2,000 sample settings, respectively, while Qwen-VL-Chat shows more substantial gains of 1.4\% and 0.4\% in the same settings, compared with the second best performance.  Notably, these improvements are observed in both dense text and scene text datasets. In contrast, given the small number of fine-tuning samples,  QA-ViT  method reveals limited improvement on dense text datasets such as KLC and WTQ, and it even performs worse than solely fine-tuning with LoRA. This is attributed to the fact that QA-ViT utilizes all query tokens for interaction with the vision block, which becomes noisy and inefficient for learning in data-scarce scenarios. To ensure a fair and consistent comparison, we compare results within each model. Other \textcolor{gray}{baselines} utilized higher resolutions and were fully trained on the datasets; we include them to provide the reader with more context regarding the current progress in solving the VDU task.
Conclusively, across different experimental settings, our method consistently enhances the visual representations of the image encoder within VLMs in text-rich scenarios, thereby bolstering the performance of VDU with the smallest additional overhead in the inference phase.

Figure \ref{fig:qualitative_result} showcases the qualitative results generated by the Qwen-VL-Chat model \cite{bai2023qwen}, comparing the version fine-tuned with  QA-ViT  \cite{ganz2024questionaware} to our proposed \SystemName\ approach across a varied collection of document images. More qualitative visualizations are provided in Supplementary Material - Section \ref{supp:Visualization}. The results clearly demonstrate that our method is not only effective in extracting information from dense text documents (as shown in the left and right images) but also significantly enhances the reasoning capabilities of the VLMs (as evident in the middle images). The integration of a single query embedding with a query agnostic module, combined with our fuse and defuse training modules, significantly improves the efficacy of vision models. This enhancement facilitates a deeper understanding in text-rich environments for the VLMs.

\subsection{Ablation Study}
\textbf{Effect of different modules.} Table \ref{tab:ablation_modules} presents the effects of each module proposed in our paper: the query-agnostic module, spherical augmentation, and entropy regularization. Additionally, we experiment with all query token embeddings. The experiments are conducted using 1,000 samples on Qwen-VL-Chat model. Notably, without the query-agnostic module, the model 
performance decreases by 0.54\% compared to the full QID model, due to the shift in vision representation caused by the incorporation of the query token as input. In the absence of noise augmentation and entropy regularization, the model typically performs worse, particularly on the KLC and VizWiz datasets, compared to the full QID model. This observation confirms that both fuse and defuse learning steps enable the model to learn from scarce data more efficiently. Lastly, using all query token embeddings (\textit{with} $\textbf{q}$) tends to limit our ability to integrate fuse and defuse learning steps, and meanwhile to introduce redundant tokens in training, resulting in a 0.38\% lower performance than our proposed \SystemName. These findings validate the necessity of our learning strategies, which incorporate query tokens into vision models,  thereby enhancing the comprehension of text-rich documents in VLMs.

\begin{table}[t!]
\centering
 \resizebox{0.489\textwidth}{!}{ 
\begin{tabular}{l|ccc|cc|c}
\hline
Settings & \textbf{InfoVQA} & \textbf{KLC} & \textbf{WTQ} & \textbf{VizWiz} & \textbf{ST-VQA} & \textbf{\textit{Avg.}} \\
&  \textcolor{gray}{\textit{ANLS $\uparrow$}} & \textcolor{gray}{\textit{F1 $\uparrow$}} & \textcolor{gray}{\textit{ACC $\uparrow$}} & \textcolor{gray}{\textit{VQA Score $\uparrow$}} & \textcolor{gray}{\textit{ANLS $\uparrow$}} \\
\hline\hline
\textit{w/o query-agnostic}& 34.60 & 27.71 & 22.96 & 39.89 & 63.46 & {37.72}          \\
\textit{w/o sinusoidal}& 34.60 & 27.31 & 22.45 & 40.97 & 63.96 & {37.85}          \\
\hline
\textit{w/o fuse step} & 33.84 & 27.44 & 23.50 & 39.02 & 63.92 & {37.54}          \\
\textit{w/o defuse step} & 34.77 & 27.60 & 23.58 & 37.94 & 62.90 & {37.36}          \\
\hline
 \textit{with} $\textbf{q}$& 33.89 & 27.63 & 23.21 & 40.39 & 64.28 & {37.88}          \\ \hline
\SystemName\ (\textit{\textbf{ours}}) & 34.18 & 27.81 & 23.56 & 42.08 & 63.69 & {\textbf{38.26}}\\
\hline\hline
 
\end{tabular}
}
\caption{\textbf{Ablation studies}. Impacts of various proposed modules on the Qwen-VL-Chat models fine-tuned with 1,000 samples.  }
\label{tab:ablation_modules}
\vspace{-10pt}
\end{table}

\textbf{Fine-tuning with extremely limited and full data.} To demonstrate our method’s effectiveness across learning conditions, we fine-tuned Qwen-VL-Chat using 500 samples per dataset and the full datasets, with consistent training settings except for a single epoch for full datasets. The results are shown in Table \ref{tsb:ablation_full_data}. In the extremely limited data scenario, with only 500 training samples, our method still achieved the best performance among all baselines, showing an average of 1.22\% improvement over  QA-ViT. Meanwhile, with full datasets, our \SystemName\ achieved the highest performance across all five datasets, obtaining an average improvement of 1.38\% compared to QA-ViT. Conversely, limitations of QA-ViT  are revealed in some scenarios when trained with full datasets, typically underperforming compared to LoRA by 0.5\%  on KLC — a dense text dataset. These results highlight the superiority of our method over QA-ViT, not only in data-scarce regimes but entire datasets, in terms of enhancing the image encoder's ability to discern more effective cues in text-rich environments.

\begin{table}[t!]
\centering
 \resizebox{0.489\textwidth}{!}{ 
\begin{tabular}{l|ccc|cc|c}
\hline
\textbf{Method} & \textbf{InfoVQA} & \textbf{KLC} & \textbf{WTQ} & \textbf{VizWiz} & \textbf{ST-VQA} & \textbf{\textit{Avg.}} \\
&  \textcolor{gray}{\textit{ANLS $\uparrow$}} & \textcolor{gray}{\textit{F1 $\uparrow$}} & \textcolor{gray}{\textit{ACC $\uparrow$}} & \textcolor{gray}{\textit{VQA Score $\uparrow$}} & \textcolor{gray}{\textit{ANLS $\uparrow$}} \\
\hline\hline
\textit{\textbf{PEFT on 500}} &&&&&&\\
\quad DoRA \cite{liu2024dora} &  34.00 & 26.02 & 22.47 & 38.00 & 57.04 & 35.51       \\
\quad LoRA \cite{hu2021lora} &    34.56    & 26.60        & 22.72       & 38.40     & 57.63       & 35.98       \\
\quad VPT \cite{jia2022visual}  &   31.00       & 24.96       & 19.34       & 42.92    & 55.76       & 34.80         \\
\quad  QA-ViT  \cite{ganz2024questionaware} &     33.24    & 25.98       & 21.60        & 39.72    & 61.82       & 36.47     \\ 
\quad \SystemName\ (\textit{\textbf{ours}}) & {34.17}    & {27.17}       & {23.85}       & {39.87}    & {63.41}       & \textbf{37.69} \\
\hline
\textit{\textbf{PEFT on Full}} &&&&&&\\
\quad DoRA \cite{liu2024dora} & 34.56 & 28.78 & 24.25 & 40.19 & 58.89 & 37.33          \\
\quad LoRA \cite{hu2021lora} & 34.55 & 29.15 & 24.27 & 40.51 & 58.50 & 37.40         \\
\quad VPT \cite{jia2022visual} & 30.37 & 27.70 & 21.06 & 39.39 & 54.86 & 34.68           \\
\quad  QA-ViT  \cite{ganz2024questionaware} & 34.95 & 28.65 & 25.19 & 39.21 & 63.21 & {38.28}          \\ 
\quad \SystemName\ (\textit{\textbf{ours}}) & {35.24} & {30.17} & {26.02} & {41.48} & {65.39} & \textbf{39.66}\\
\hline\hline
 
\end{tabular}
}
\caption{\textbf{Ablation studies}. Performance of our proposed method when integrated with the Qwen-VL-Chat model, trained on a very small number of tuning samples (500) as well as on the full dataset.}
\label{tsb:ablation_full_data}
\vspace{-10pt}
\end{table}
\section{Limitation \& Future Work}
While our proposed QID method demonstrates superior performance across various datasets, we acknowledge two primary limitations that warrant further investigation. First, our approach depends on an external pre-trained CLIP model \cite{li2023clip}, constrained to 77 tokens, potentially limiting its ability to handle longer or more complex queries; we plan to mitigate this by exploring state-of-the-art solutions like Long-CLIP \cite{zhang2024long} and its variants to support extended text inputs. Secondly, our current experiments focus on single-hop QA tasks, whereas real-world VDU often involves multi-hop QA requiring reasoning across multiple document regions. Although our design—integrating query-aware and query-agnostic modules into the final ViT layers (Fig. \ref{fig:main_diagram})—efficiently avoids multiple forward passes per query, its effectiveness for multi-hop reasoning remains untested; we intend to investigate adaptive query propagation mechanisms to address this.

\section{Conclusion}
\label{sec:conclusion}
This paper presents \textbf{\SystemName}, a novel fine-tuning approach of enhancing OCR-free Visual Document Understanding (VDU) for Vision-Language Models (VLMs) in data-scarce regimes. By integrating a single query vector into the vision encoder without modifying its core architecture, QID effectively directs attention to query-relevant visual regions in a computationally efficient manner. Our unique fuse and defuse learning steps strengthen the query representation in data-scarce settings, while our query-agnostic module ensures robust positional encoding, supporting the model's adaptability to various document layouts. Our experimental results demonstrate that VLMs equipped with \SystemName\ achieve evident performance gains across various datasets compared to baseline models, particularly excelling for dense-text tasks, with minimal overhead. Future directions include exploring multi-turn query representations to enable interactive VDU and scaling QID for larger, more complex models in diverse application scenarios.


\bibliographystyle{ieee_fullname}
\bibliography{egbib}

\begin{appendices}
\clearpage
\nocitesec{*}
 
\section{Fine-tuning Datasets}
\label{supp:supp_data}
During our fine-tuning stage, VLMs are optimized using a variety of datasets with different tasks. These datasets, introduced in Table \ref{tab:dataset}
, include InfographicVQA \cite{mathew2022infographicvqa}, Kleister Charity \cite{stanislawek2021kleister}, WikiTableQuestions \cite{pasupat2015compositional}, VizWiz-VQA \cite{gurari2018vizwiz}, and ST-VQA \cite{biten2019scene}, and are briefly described as follows:

\textbf{InfographicVQA (InfoVQA)} \cite{mathew2022infographicvqa}: This dataset is a collection of over five thousand infographic images, along with a large number of question-answer pairs. These infographics are sourced from various web domains and feature diverse layouts and designs. The InfographicVQA challenges vision language models to interpret and reason over complex visual documents, often necessitating understanding of graphical elements, data visualization, reasoning, and arithmetic skills.

\textbf{Kleister Charity (KLC)} \cite{stanislawek2021kleister}: This dataset consists of annual financial reports from UK charity organizations. The task involves key information extraction (KIE) such as charity names, addresses, charity numbers, and reporting dates. Primarily comprising scanned documents, this dataset poses challenges due to its length, diverse layout, and the necessity to interpret both text and structural features.

\textbf{WikiTableQuestions (WTQ)} \cite{pasupat2015compositional}: This dataset includes question and answer pairs collected from thousands of HTML tables extracted from Wikipedia. The questions are designed to be complex, requiring multi-step reasoning and various data operations such as comparison, aggregation, and arithmetic computation.

\textbf{VizWiz-VQA (VizWiz)} \cite{gurari2018vizwiz}: Comprising a large number of question-answer pairs, this dataset features images captured by blind individuals using mobile phones and spoken questions about those images. Unique in terms of its image quality, which is often blurred, and the nature of its questions, a significant portion of the images are unanswerable due to poor image quality.

\textbf{ST-VQA} \cite{biten2019scene}: Designed specifically for understanding textual information within natural images, this dataset requires models to read and interpret scene text to accurately answer questions. It includes a large collection of images sourced from various public datasets such as COCO-text, Visual Genome, and ImageNet, challenging models to comprehend images across a wide range of scenarios and textual appearances within images.

\section{More Qualitative Results}
\label{supp:Visualization}

\begin{figure*}[t!]
\centering
\includegraphics[width=1\textwidth]{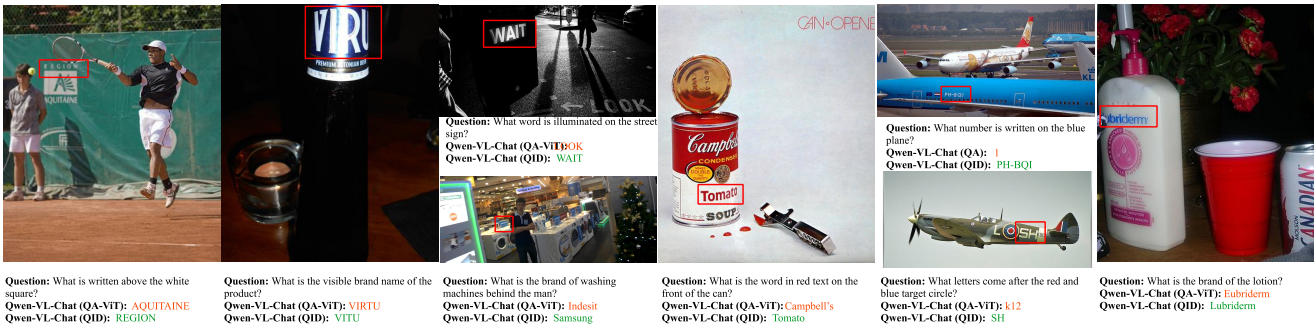}
\caption{More qualitative results between between  QA-ViT  and our \SystemName\ with Qwen-VL-Chat model. Image regions with answers are highlighted.}
\label{fig:more_qualitative_result}
\end{figure*}

Figure \ref{fig:more_qualitative_result} presents additional quantitative results derived from various evaluation datasets comparing QA-ViT method \cite{ganz2024questionaware} and our \SystemName\ method, implemented with Qwen-VL-Chat \cite{bai2023qwen}. These results highlight the effectiveness of our method in enhancing the vision model’s ability to identify relevant visual cues and improve comprehension in both text-rich and natural scene environments.

Furthermore, we outline the limitations of our approach in Figure \ref{fig:fail_case}. Although our method aids in enhancing understanding of text-rich images, it does not significantly improve the model's reasoning and arithmetic capabilities. Consequently, our future research will focus on refining the model's ability to perform complex reasoning tasks more effectively in dense-text settings.

\section{Broader Impact}
\label{supp:impact}
 The enhanced capabilities of vision-language models (VLMs) offer substantial promise for improving document comprehension in environments with extensive textual content. However, the interaction between question embeddings and vision representations remains relatively unexplored. Our approach encourages this interaction with limited fine-tuning samples while preserving the structural integrity of pre-trained VLMs. It also minimizes the necessity for extensive retraining, thereby reducing the computational resources required for deploying sophisticated AI solutions. Additionally, our method's efficiency with limited data can decrease both the time and costs involved in annotating large datasets, enhancing the accessibility and affordability of advanced document understanding technologies. We encourage the research community to further explore and adopt our QID for text-intensive tasks, anticipating significant benefits in various applications.

 \begin{figure*}[t!]
\centering
\includegraphics[width=1\textwidth]{Figures/fail_case.pdf}
\caption{Failure cases of \SystemName\ on documents and questions require arithmetic and reasoning skills. Image regions with answers are highlighted. }
\label{fig:fail_case}
\end{figure*}

\end{appendices}
\end{document}